\documentclass{article}

\usepackage[preprint]{neurips_2023}

\usepackage{natbib}
\setcitestyle{numbers,square}

\usepackage[utf8]{inputenc} % allow utf-8 input
\usepackage[T1]{fontenc}    % use 8-bit T1 fonts
\usepackage{hyperref}       % hyperlinks
\usepackage{url}            % simple URL typesetting
\usepackage{booktabs}       % professional-quality tables
\usepackage{amsfonts}       % blackboard math symbols
\usepackage{nicefrac}       % compact symbols for 1/2, etc.
\usepackage{microtype}      % microtypography
\usepackage{xcolor}         % colors

\usepackage{times}
\usepackage{epsfig}
\usepackage{graphicx}
\usepackage{amsmath}
\usepackage{amssymb}

\usepackage{amsmath}
\usepackage{amssymb}
\usepackage{mathtools}
\usepackage{amsthm}

\usepackage{multirow}

\usepackage{algorithm}
\usepackage[]{algpseudocode}

\title{Lookahead Drifting Model}

\author{%
  Guoqiang Zhang  \\
  University of Exeter \\
  \texttt{g.z.zhang@exeter.ac.uk} \\
     \And  Kenta Niwa \\
    NTT Communication Science Laboratories \\
   \texttt{kenta.niwa@ntt.com}  \\
   \And
   W. Bastiaan Kleijn \\ 
Victoria University of Wellington \\ \texttt{bastiaan.kleijn@vuw.ac.nz} \\
}

\begin{document}

\maketitle

\begin{abstract}
Recently, a new paradigm named \emph{drifting model} has been proposed for mapping distributions, which achieves the SOTA image generation performance over ImageNet via one-step neural functional evaluation (NFE). The basic idea is to compute a drifting term at each training iteration and then push the output of the model towards the direction of the drifting term. In this paper, we propose a \emph{lookahead drifting model}. At each training iteration, we compute a set of drifting terms sequentially. Each drifting term is calculated by making use of previously computed ones as well as the positive samples and the output of the model. %One key step is to properly scale the drifting terms so that their magnitudes are in a comparable range. 
In principle, the drifting terms obtained at a later stage capture higher order gradient information towards the positive samples. At each training iteration, the model is optimized by pushing its output towards the direction of the (weighted) summation of the drifting terms. Experimental results on toy examples and CIFAR10 demonstrate the better performance of the new method than the baseline using 1-step NFE.

%The abstract paragraph should be indented \nicefrac{1}{2}~inch (3~picas) on both the left- and right-hand margins. Use 10~point type, with a vertical spacing (leading) of 11~points. The word \textbf{Abstract} must be centered, bold, and in point size 12. Two line spaces precede the abstract. The abstract must be limited to one paragraph.
\end{abstract}

\section{Introduction}

Mapping between distributions for the generation purpose has made significant progress in the last two decades. Various learning paradigms have been proposed by following different principles. For instance, the variational autoencoder (VAE) attempts to train the probabilistic decoder via the help of the encoder under the assumption (typically) of Gaussian distribution for the latent random vectors  \citep{Kingma_VAE}. Generative adversarial networks (GANs) introduce an additional discriminator to guide the learning process of the generator \cite{Arjovsky17WGAN, Goodfellow14GAN}. Upon convergence, the generator produces high quality samples that are indistinguishable from real data samples, and the discriminator cannot provide useful gradient information anymore.  While the GAN framework is theoretically grounded, its training process is known to be unstable.

To avoid the training ditfficulty of GAN, normalizing flows (NFs) have been proposed to gradually transform one distribution to another via a sequence of reversible neural blocks \cite{Kobyzev20NormFlow}. Various methods have been proposed on designing effective reversible blocks \cite{Kingma18Glow, Dinh14Nice, Dinh16DensityEsti, Gomez17ReverseResnet, Jacobsen2018IREVENT, Behrmann19InvResNet}. One practical issue with NFs is that in general, the generation quality cannot match that of GAN, which is probably due to the architectural constraints of reversibility. 

In the last few years, diffusion and flow-matching models \cite{Ho20DDPM, Ricky24FlowMatching} have achieved great success for mapping distributions. Inspired by NFs, both approaches introduce a timestep $t$ as an additional input to a neural network such that as $t$ increases from the starting point to the endpoint, the distribution of the output of the neural network smoothly transforms one distribution to another. Once the neural network is well trained, high quality data samples can be generated by solving either a stochastic differential equation (SDE) or an ordinary differential equations (ODE) via a certain NFE (number of function evaluations), see for example \cite{Song21DDIM, Karras22EDM, Lu22DPM_SolverPlus, Bao22DPM, Zhang24BDIA} for different SDE/ODE solvers.               

Recently, the drifting model \cite{Deng26drifting} was proposed with the aim to enable sampling with only a single function evaluation (NFE=1). In essence, the new method brings the sampling iterations of diffusion (or flow-matching) models into the training process. At each training iteration, a drifting term is computed, and the model output is pushed slightly in its direction.  Upon convergence, the drifting term approaches to zero, indicating that the distribution of the model output coincides with the data distribution. In other words, the drifting term provides (weak) gradient information to properly guide the update of the model's output toward real data distribution iteratively. The research works in \cite{Lai26drifting} identified the relation between drifting model and earlier methods such as score-matching and maximum-mean discrepancy (MMD).

In this paper, we attempt to extend drifting model by computing and exploring multiple drifting terms sequentially per training iteration. Each drifting term is computed by using the previously computed ones as well as the model's output and the real data samples. The main purpose for doing so is to capture higher order gradient information embedded in the model output and real data samples that is missing in the standard drifting model. Similarly, upon convergence, all the drifting terms approach zero, indicating that the data distribution has been well matched. Experiments on toy examples and CIFAR10 show that the new method produces better performance than the standard drifting model.

\section{Preliminary}
\label{sec:pre}
As mentioned earlier, the diffusion and flow matching models have to generate new samples by solving an SDE or ODE in an iterative manner, which is time-consuming. In contrast, drifting models attempt to move the sampling iterations to the training process. Mathematically, we denote the mapping function as $f_{\boldsymbol{\theta}}(\boldsymbol{\epsilon})$ where $\boldsymbol{\epsilon}\sim \mathcal{N}(\boldsymbol{0}, \boldsymbol{I})$. With increasing training iteration number, the method aims to gradually push the generated distribution $q(f_{\boldsymbol{\theta}}(\boldsymbol{\epsilon}))$ towards that of the data distribution $p(\cdot)$ with the help of the so-called \emph{drifting term}. In brief, drifting models inherit the advantages of GANs and diffusion/flow-matching models in a unified manner.  

At each training iteration, a drifting term $\mathbf{V}_{p,q}(f_{\boldsymbol{\theta}}(\boldsymbol{\epsilon}))$ is firstly constructed based on the data distribution $p$ and the generated distribution $q$ (which will be explained later on). The mapping function $f_{\boldsymbol{\theta}}(\boldsymbol{\epsilon})$ is then optimized by minimizing  
\begin{align}
\arg\min_{\boldsymbol{\theta
}} \Bigg(L_{\boldsymbol{\theta
}} \equiv \mathbb{E}_{\boldsymbol{\epsilon}\sim \mathcal{N}(\boldsymbol{0},\boldsymbol{I})} \Big[ \big\| f_{\boldsymbol{\theta}}(\boldsymbol{\epsilon}) - \text{stopgrad}\big(\underbrace{f_{\boldsymbol{\theta}}(\boldsymbol{\epsilon}) + \mathbf{V}_{p,q}(f_{\boldsymbol{\theta}}(\boldsymbol{\epsilon})}_{ f^1_{\boldsymbol{\theta}}(\boldsymbol{\epsilon}) })\big) \big\|^2 \Big]\Bigg),
\label{equ:drifting_loss}
\end{align}
where ``$\text{stopgrad}$" ensures that the second function $f_{\boldsymbol{\theta}}(\boldsymbol{\epsilon})$ is frozen when updating the model.  The drifting term makes the distribution of the target $f^1_{\boldsymbol{\theta}}(\boldsymbol{\epsilon})$ a bit closer to $p(\cdot)$. 

The work \cite{Deng26drifting} proposes to compute the drifting term in the following form:
\begin{align}
\mathbf{V}_{p,q}(\mathbf{x}) := \mathbf{V}_p^{+}(\mathbf{x}) - \mathbf{V}_q^{-}(\mathbf{x}),
\label{equ:drift}
\end{align}
where $\mathbf{V}_p^{+}(\mathbf{x})$ and $\mathbf{V}_p^{-}(\mathbf{x})$ denote attraction and repulsion forces, respectively, which are given by
\begin{align}
&\mathbf{V}_p^{+}(\mathbf{x}) := \frac{1}{\mathbb{E}_{\mathbf{y}^{+} \sim p} \big[ k(\mathbf{x}, \mathbf{y}^{+}) \big]} \, \mathbb{E}_{\mathbf{y}^{+} \sim p} \big[ k(\mathbf{x}, \mathbf{y}^{+}) (\mathbf{y}^{+} - \mathbf{x}) \big] \label{equ:drift_pos} \\ 
&\mathbf{V}_q^{-}(\mathbf{x}) := \frac{1}{\mathbb{E}_{\mathbf{y}^{-} \sim q} \big[ k(\mathbf{x}, \mathbf{y}^{-}) \big]} \, \mathbb{E}_{\mathbf{y}^{-} \sim q} \big[ k(\mathbf{x}, \mathbf{y}^{-}) (\mathbf{y}^{-} - \mathbf{x}) \big], \label{equ:drift_neg}
\end{align}
where $k(\mathbf{x},\mathbf{y})=e^{\|\mathbf{x}-\mathbf{y}\|_2/\tau}$ is the Laplace kernel with a temperature  parameter $\tau$. % for controlling the strength of the contributions .

One essential property with the expression (\ref{equ:drift}) is that it is anti-symmetric: 
\begin{align}
\mathbf{V}_{p,q}(\mathbf{x}) = -\mathbf{V}_{q,p}(\mathbf{x}).\label{equ:antisym}
\end{align}
The above property ensures that when $q=p$ (i.e., the generated distribution matches with the data distribution), the drifting term satisfies $\mathbf{V}_{p,q}(\mathbf{x})=0$ and the model's output reaches the stationary point. That is, once the model converges, the model's output would not deviate easily from the stationary point.

%Substituting into $\mathbf{V}_{p,q} = \mathbf{V}_p^{+} - \mathbf{V}_q^{-}$, we obtain the compact form
%$$
%\mathbf{V}_{p,q}(\mathbf{x}) = \frac{1}{Z_p(\mathbf{x}) Z_q(\mathbf{x})} \, \mathbb{E}_{\mathbf{y}^{+} \sim p, \, \mathbf{y}^{-} \sim q} \big[ k(\mathbf{x}, \mathbf{y}^{+}) \, k(\mathbf{x}, \mathbf{y}^{-}) \, (\mathbf{y}^{+} - \mathbf{y}^{-}) \big].
%$$

\section{Lookahead Drifting Models}
In this section, we first motivate the benefits for computing multiple drifting terms per training iteration.  We then present and analyze the update expressions of lookahead drifting models.  

\subsection{Motivation} We note from Section~\ref{sec:pre} that in principle, the distribution $q_1(\cdot)$ of the target $f^1_{\boldsymbol{\theta}}(\boldsymbol{\epsilon})$ is closer to the data distribution $p(\cdot)$ than $q(\cdot)$ of the current prediction $f_{\boldsymbol{\theta}}(\boldsymbol{\epsilon})$. Therefore, by iteratively pushing $f_{\boldsymbol{\theta}}(\boldsymbol{\epsilon})$ to be closer to the target $f^1_{\boldsymbol{\theta}}(\boldsymbol{\epsilon})$, 
the model learns to match the data distribution.

One natural extension of standard drifting model is to compute and explore an additional drifting term $\mathbf{V}_{p,q_1}(f^1_{\boldsymbol{\theta}}(\boldsymbol{\epsilon}))$ for $f^1_{\boldsymbol{\theta}}(\boldsymbol{\epsilon})$ when updating the model parameter $\boldsymbol{\theta}$. The additional drifting would in principle provide higher order gradient information between the two distributions $p(\cdot)$ and $q(\cdot)$. Following the above reasoning, one can compute multiple drifting terms 
$\{\mathbf{V}_{p,q_i}(f^i_{\boldsymbol{\theta}}(\boldsymbol{\epsilon}))\}_{i=0}^k$ sequentially in preparation for updating the model, where for the case of $i=0$, $f^0_{\boldsymbol{\theta}}(\boldsymbol{\epsilon})=f_{\boldsymbol{\theta}}(\boldsymbol{\epsilon})$. We use $q_{i}(\cdot)$ to represent the distribution for the prediction $f^{i}_{\boldsymbol{\theta}}(\boldsymbol{\epsilon})$.         

In brief, while the standard drifting model computes a single drifting term via the distribution pair $(q(\cdot), p(\cdot))$, we propose to compute multiple drifting terms via multiple distribution pairs 
$\{(q_i(\cdot), p(\cdot)) | k\geq i\geq 0  \}$, where in the ideal case, $q_i(\cdot)$ would be closer to the data distribution $p(\cdot)$ than its preceding one $q_{i-1}(\cdot)$. We use the term "lookahead" to indicate that we exploit those additional distributions between $q(\cdot)$ and $p(\cdot)$ when updating the model.   

\subsection{Update expressions}
Based on our earlier analysis, we propose to update the model parameter $\boldsymbol{\theta}$ by utilizing a set of drifting terms $\{\mathbf{V}_{p,q_i}(f^i_{\boldsymbol{\theta}}(\boldsymbol{\epsilon}))\}_{i=0}^k$ in the following form:
\begin{align}
\arg\min_{\boldsymbol{\theta
}} \Bigg(L_{\boldsymbol{\theta
}} = \mathbb{E}_{\boldsymbol{\epsilon}\sim \mathcal{N}(\boldsymbol{0},\boldsymbol{I})} \Big[ \big\| f_{\boldsymbol{\theta}}(\boldsymbol{\epsilon}) - \text{stopgrad}\big({f_{\boldsymbol{\theta}}(\boldsymbol{\epsilon}) + \sum_{i=0}^k \mathbf{V}_{p,q_i}(f^i_{\boldsymbol{\theta}}(\boldsymbol{\epsilon})})\big) \big\|^2 \Big]\Bigg),
\label{equ:drifting_loss_ahead}
\end{align}
where for each $i\geq 1$, $f^i_{\boldsymbol{\theta}}(\boldsymbol{\epsilon})$ is computed to be
\begin{align}
f^i_{\boldsymbol{\theta}}(\boldsymbol{\epsilon})=f^{i-1}_{\boldsymbol{\theta}}(\boldsymbol{\epsilon}) + \mathbf{V}_{p,q_{i-1}}(f^{i-1}_{\boldsymbol{\theta}}(\boldsymbol{\epsilon})),
\label{equ:dis_bypass}
\end{align}
where the expression for $\mathbf{V}_{p,q_{i-1}}(f^{i-1}_{\boldsymbol{\theta}}(\boldsymbol{\epsilon}))$ follows directly from (\ref{equ:drift})-(\ref{equ:drift_neg}). We notice that the mapping from $f_{\boldsymbol{\theta}}^{i-1}(\cdot)$ to $f_{\boldsymbol{\theta}}^{i}(\cdot)$ is deterministic. This indicates that the distribution $q_i(\cdot)$ of $f_{\boldsymbol{\theta}}^{i}(\cdot)$ is fully determined by $q_0(\cdot)=q(\cdot)$ and $p(\cdot)$.

\textbf{Similarity with ResNet-type of models}: In brief, ResNet-type (e.g., ResNet, Transformer)  models introduce a skip-connection every neural block to allow gradients to flow directly backward through shortcut paths, bypassing blocks to prevent signal loss. Similarly, the expression for $f^i_{\boldsymbol{\theta}} (\boldsymbol{\epsilon})$ in (\ref{equ:dis_bypass}) can be viewed as being computed by a skip-connection. The final expression $f^k_{\boldsymbol{\theta}} (\boldsymbol{\epsilon})$ is obtained by stacking $k$ drifting terms via consecutive skip-connections. 

It is known that the neural blocks at a later stage in ResNet-type models attempt to capture high-level semantic information of the data samples. Similarly, the drifting terms at a later stage are introduced to measure high-order differences between the data distribution $p(\cdot)$ and the generated distribution $q(\cdot)$.          

\textbf{Anti-symmetric property}: Based on the definition (\ref{equ:drift})-(\ref{equ:drift_neg}), it is clear that the anti-symmetric property holds for each drifting term $\mathbf{V}_{p,q_{i}}(f^{i}_{\boldsymbol{\theta}}(\boldsymbol{\epsilon}))$, $k\geq i\geq 0$. It is known from (\ref{equ:antisym}) that $q=p$ leads to the convergence condition $ \mathbf{V}_{p,q_{0}}(f^{0}_{\boldsymbol{\theta}}(\boldsymbol{\epsilon}))=0$. As a result, for any $i>0$, we have $q_i=p$ and $ \mathbf{V}_{p,q_{i}}(f^{i}_{\boldsymbol{\theta}}(\boldsymbol{\epsilon}))=0$.

\textbf{High order gradient information}: Without loss of generality, we study the impact of $\mathbf{V}_{p,q_1}(f^1_{\boldsymbol{\theta}}(\boldsymbol{\epsilon}))$ for the special case of $k=1$. Our first objective is show that the drifting term of $k=1$ is not a simple replication of the drifting term of $k=0$. Secondly, we argue that $\mathbf{V}_{p,q_1}(\cdot)$ indeed captures the difference between $p$ and $q$ that is missing in $\mathbf{V}_{p,q}(\cdot)$. 

The quantity $\mathbf{V}_{p,q_1}(\cdot)$ can be represented as
\begin{align}
& \mathbf{V}_{p,q_1}(f^1_{\boldsymbol{\theta}}(\boldsymbol{\epsilon})) \nonumber\\
%&=\mathbf{V}_{p,q_1}\Big(f_{\boldsymbol{\theta}}(\boldsymbol{\epsilon}) + \mathbf{V}_{p,q}(f_{\boldsymbol{\theta}}(\boldsymbol{\epsilon}))  \Big) \nonumber \\
%&=  \mathbf{V}_{p,q_1}\Big(f_{\boldsymbol{\theta}}(\boldsymbol{\epsilon}) + \mathbf{V}_p^{+}(f_{\boldsymbol{\theta}}(\boldsymbol{\epsilon})) - \mathbf{V}_q^{-}(f_{\boldsymbol{\theta}}(\boldsymbol{\epsilon})) \Big) \nonumber \\
&=  \mathbf{V}_{p}^{+}\Big(f^1_{\boldsymbol{\theta}}(\boldsymbol{\epsilon}) \Big) - \mathbf{V}_{q_1}^{-}\Big(f^1_{\boldsymbol{\theta}}(\boldsymbol{\epsilon}) \Big)  \nonumber \\
%&= \mathbf{V}_{p}^{+}\Big(f_{\boldsymbol{\theta}}(\boldsymbol{\epsilon}) + \mathbf{V}_p^{+}(f_{\boldsymbol{\theta}}(\boldsymbol{\epsilon})) - \mathbf{V}_q^{-}(f_{\boldsymbol{\theta}}(\boldsymbol{\epsilon})) \Big) - \mathbf{V}_{q_1}^{-}\Big(f_{\boldsymbol{\theta}}(\boldsymbol{\epsilon}) + \mathbf{V}_p^{+}(f_{\boldsymbol{\theta}}(\boldsymbol{\epsilon})) - \mathbf{V}_q^{-}(f_{\boldsymbol{\theta}}(\boldsymbol{\epsilon})) \Big)  \nonumber \\
&\stackrel{(a)}{=} \, \frac{1}{Z_1^{+}} \mathbb{E}_{\mathbf{y}^{+} \sim p} \big[ k
\left(f^1_{\boldsymbol{\theta}}(\boldsymbol{\epsilon}), \mathbf{y}^{+}\right) \mathbf{y}^{+} \big]  - \frac{1}{Z_1^{-}}\mathbb{E}_{\mathbf{y}^{-} \sim q_1} \big[ k\left(f^1_{\boldsymbol{\theta}}(\boldsymbol{\epsilon}), \mathbf{y}^{-}\right) \mathbf{y}^{-} \big] \nonumber \\
& \stackrel{(b)}{=} \, \frac{1}{Z_1^{+}} \mathbb{E}_{\mathbf{y}^{+} \sim p} \big[ k
\left(f^1_{\boldsymbol{\theta}}(\boldsymbol{\epsilon}), \mathbf{y}^{+}\right) \mathbf{y}^{+} \big]  \nonumber \\
& \hspace{5mm} - \frac{1}{Z_1^{-}}\mathbb{E}_{\mathbf{y}^{-} \sim q} \big[ k\left(f^1_{\boldsymbol{\theta}}(\boldsymbol{\epsilon}), \mathbf{y}^{-} + \mathbf{V}_p^{+}(\mathbf{y}^{-}) - \mathbf{V}_q^{-}(\mathbf{y}^{-})  \right) (\mathbf{y}^{-}+\mathbf{V}_p^{+}(\boldsymbol{y}^{-}) - \mathbf{V}_q^{-}(\boldsymbol{y}^{-})) \big] \label{equ:V_1} 
\end{align}
where in step $(a)$, the two constants $Z_1^{+}$ and $Z_1^{-}$ are given by 
\begin{align}
Z_1^{+}&={\mathbb{E}_{\mathbf{y}^{+} \sim p} \big[ k(f^1_{\boldsymbol{\theta}}(\boldsymbol{\epsilon}), \mathbf{y}^{+}) \big]} \nonumber \\
Z_1^{-}&={\mathbb{E}_{\mathbf{y}^{-} \sim q_1} \big[ k(f^1_{\boldsymbol{\theta}}(\boldsymbol{\epsilon}), \mathbf{y}^{-}) \big]}. \nonumber
\end{align}
Step $(b)$ is derived by using the fact for any $\mathbf{y}^{-}\sim q(\cdot)$, $\mathbf{y}^{-}+\mathbf{V}_p^{+}(\mathbf{y}^{-})-\mathbf{V}_q^{-}(\mathbf{y}^{-})\sim q_1(\cdot)$.  %the distribution for the 2nd term is changed from $q_1(\cdot)$ to $q(\cdot)$.

Next we investigate the difference between $\mathbf{V}_{p,q_1}(\cdot)$ and $\mathbf{V}_{p,q}(\cdot)$. One can easily derive the expression for $\mathbf{V}_{p,q}(\cdot)$, which is given by 
\begin{align} 
\mathbf{V}_{p,q}(f_{\boldsymbol{\theta}}(\boldsymbol{\epsilon}))
& = \frac{1}{Z_0^{+}} \mathbb{E}_{\mathbf{y}^{+} \sim p} \big[ k
\left(f_{\boldsymbol{\theta}}(\boldsymbol{\epsilon}), \mathbf{y}^{+}\right) \mathbf{y}^{+} \big]   - \frac{1}{Z_0^{-}}\mathbb{E}_{\mathbf{y}^{-} \sim q} \big[ k\left(f_{\boldsymbol{\theta}}(\boldsymbol{\epsilon}), \mathbf{y}^{-} \right) \mathbf{y}^{-} \big], \label{equ:V_0}
\end{align}
where $(Z_0^{+},Z_0^{-})$ are computed in a similar manner as $(Z_1^{+},Z_1^{-})$. 
Suppose $p\neq q$. For any sample $\mathbf{y}^{-}\sim q(\cdot)$, we have $\mathbf{V}_p^{+}(\mathbf{y}^{-})\neq \mathbf{V}_q^{-}(\mathbf{y}^{-})$. As a result, $f_{\boldsymbol{\theta}}^1(\boldsymbol{\epsilon})\neq f_{\boldsymbol{\theta}}(\boldsymbol{\epsilon})$ according to the definition (\ref{equ:dis_bypass}). By investigation of (\ref{equ:V_1}) and (\ref{equ:V_0}), the two drifting terms are not identical due to different kernel values. Interestingly, $\mathbf{V}_{p,q_1}$ has an extra term 
\begin{align}  \frac{1}{Z_1^{-}}\mathbb{E}_{\mathbf{y}^{-} \sim q} \big[ k\left(f^1_{\boldsymbol{\theta}}(\boldsymbol{\epsilon}), \mathbf{y}^{-} + \mathbf{V}_p^{+}(\mathbf{y}^{-}) - \mathbf{V}_q^{-}(\mathbf{y}^{-})  \right) (\mathbf{V}_p^{+}(\boldsymbol{y}^{-}) - \mathbf{V}_q^{-}(\boldsymbol{y}^{-})) \big],
\label{equ:extra1}
\end{align}
which is missing in (\ref{equ:V_0}). For any pair of samples $\boldsymbol{y}_1,\boldsymbol{y}_2\in \mathbf{R}^d$, the extra term essentially considers the impact of the drifting term $\mathbf{V}_{p,q}(\boldsymbol{y}_1)$ at location $\boldsymbol{y}_1$ when computing the drifting at location $\boldsymbol{y}_2$ for $i=1$. %Intuitively speaking, if we treat $\mathbf{V}_{p,q}$ as the gradient vector. The computation for $\mathbf{V}_{p,q}$  the drifting term of $i=1$ .  
On the other hand, the standard drifting model does not explicitly consider the impact of the drifting term at one location on other locations in the sample space. In brief, the extra term (\ref{equ:extra1}) indeed captures high order difference between $p(\cdot)$ and $q(
\cdot)$. 

\section{Experimental Results}

We tested lookahead drifting method by considering both a toy example and image generation over CIFAR10. The standard drift method is taken as the baseline for comparison. In brief, the new training method produces either comparable or better performance than the baseline.

\subsection{Toy Examples}
\begin{figure}
     \centering
\vspace{0mm}\includegraphics[width=140mm]{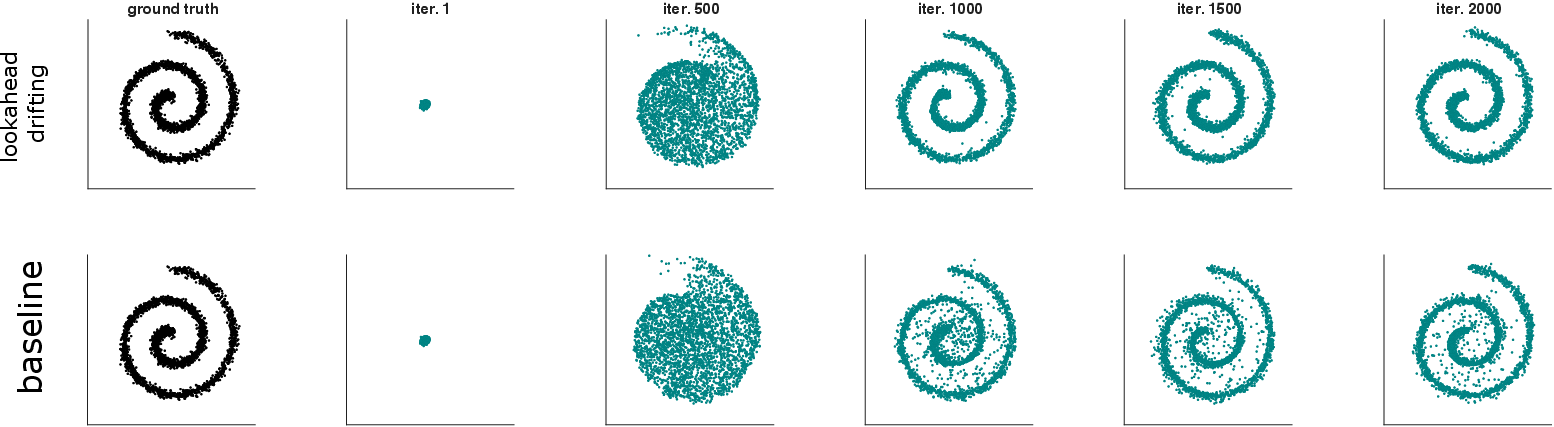}
     \caption{\small Qualitative comparison of the standard drifting and lookehead drifting with $k=3$.   }
\label{fig:toy_compare}
     \end{figure}

In the first experiment, we implemented the lookahead drifting method based on the source code\footnote{https://lambertae.github.io/projects/drifting/}. The neural architecture and learning rate of the optimizer follow directly from the original code. It is clear from Fig. \ref{fig:toy_compare} that the new training method produces competitive results as the baseline.  

\subsection{Model training over CIFAR10}

In the second experiment, we adopted the open-source\footnote{https://github.com/infatoshi/driftin} for training generative models over CIFAR10. DINOv3 was utilized as the encoder to help the training process for both the drifting and lookahead drifting methods.  The learning rates follow directly from the open-source. 

Table~\ref{tab:cifar10_fid} summarizes the FID scores of the two training methods over different training iterations. Each FID score is computed by generating and utilizing 50K images. It is clear that the lookahead drifting method produces noticeably lower FID scores than the drifting method when using a single GPU.  
 
\begin{table}[h]
\centering
\begin{tabular}{|c|c|c|c||c|}
\cline{2-5}
\multicolumn{1}{c|}{} & \multicolumn{3}{c||}{single GPU} & 8 GPU
\\ \hline
\textrm{Batchsize per GPU} & 384 & 384  & 384 & 200   \\\hline 
\textrm{Training iteration } & 30K & 40K  & 50K & 50K   \\\hline \hline
drifting \cite{Deng26drifting} & 30.15 & 29.65  &   29.67 & 19.39 \\
\hline
lookahead drifting (k=1) & \textbf{17.43} & \textbf{17.12}  & \textbf{18.81} &   \\
\hline
\end{tabular}
\caption{FID scores across training iterations. The experiment exploits encoder of DINOv3 multi-resolution.}
\label{tab:cifar10_fid}
\end{table}

\section{Conclusions}
In this paper, we extend the standard drifting model by introducing multiple drifting terms in the objective function per training iteration. The set of drifting terms are computed sequentially in order to capture high order difference between the generated distribution and data distribution. Preliminary experimental results confirm the effectiveness of those high order drifting terms in traing the generative model.

%{\small
%\bibliographystyle{abbrv}
%\bibliography{sigProcessing, sigProcessing_2nd}
%}

%%%%%%%%%%%%%%%%%%%%%%%%%%%%%%%%%%%%%%%%%%%%%%%%%%%%%%%%%%%%

\newpage

\appendix

\end{document}